\documentclass[10pt,twocolumn,letterpaper]{article}

\usepackage{wacv}
\usepackage{times}
\usepackage{epsfig}
\usepackage{graphicx}
\usepackage{amsmath}
\usepackage{amssymb}

\wacvfinalcopy % *** Uncomment this line for the final submission

\def\wacvPaperID{****} % *** Enter the wacv Paper ID here

\ifwacvfinal\fi
\usepackage{authblk}
\title{ Facial Expression Recognition Using Human to Animated-Character Expression Translation}
\author[1]{Kamran Ali\thanks{Corresponding Author: kamran@knights.ucf.edu}}
\author[2]{Ilkin Isler\thanks{ilkinsevgi@hacettepe.edu.tr}}
\author[1]{Charles Hughes\thanks{ceh@cs.ucf.edu}}
\affil[1]{Synthetic Reality Lab, Department of Computer Science University of Central Florida Orlando, Florida}
\affil[2]{Hacettepe University, Turkey}
\date{}                     %% if you don't need date to appear
\begin{document}

%%%%%%%%% TITLE

% Authors at the same institution
%\author{First Author \hspace{2cm} Second Author \\
% Authors at different institutions
%\author{Kamran Ali \hspace{2cm} Charles Hughes \\
%University of Central Florida\\
%{\tt\small kamran@knights.ucf.edu \hspace{2cm} ceh@cs.ucf.edu}
%}

% ***************************************

\maketitle
\ifwacvfinal\thispagestyle{empty}\fi

%%%%%%%%% ABSTRACT
\begin{abstract}
   Facial expression recognition is a challenging task due to two major problems: the presence of inter-subject variations in facial expression recognition dataset and impure expressions posed by human subjects. In this paper we present a novel Human-to-Animation conditional Generative Adversarial Network (HA-GAN) to overcome these two problems by using many (human faces) to one (animated face) mapping. Specifically, for any given input human expression image, our HA-GAN transfers the expression information from the input image to a fixed animated identity. Stylized animated characters from the Facial Expression Research Group-Database (FERGDB) are used for the generation of fixed identity. By learning this many-to-one identity mapping function using our proposed HA-GAN, the effect of inter-subject variations can be reduced in Facial Expression Recognition(FER). We also argue that the expressions in the generated animated images are pure expressions and since FER is performed on these generated images, the performance of facial expression recognition is improved. Our initial experimental results on the state-of-the-art datasets show that facial expression recognition carried out on the generated animated images  using our HA-GAN framework outperforms the baseline deep neural network and produces comparable or even better results than the state-of-the-art methods for facial expression recognition. 
\end{abstract}

%%%%%%%%% BODY TEXT
\section{Introduction}

Facial expression recognition (FER) plays a vital role in many world affairs and it has many interesting and exciting applications such as in human-computer interaction, intelligent tutoring system (ITS),  detecting depression \cite{r35}, interactive games, and intelligent transportation. As a consequence, FER has been a widely studied research area in the computer vision community for many decades. The recent success of deep Convolutional Neural Network (CNNs) in areas like object recognition using images and videos encourged many researchers to leverage CNNs to capture discriminative representations for facial expression recognition. Although the performance of FER has been improved quite significantly using deep neural networks, there exist many challenging problems to be addressed in order to further improve the recognition accuracy. One of the major issues with FER is that the state-of-the-art facial expression recognition datasets are small, and thus during training deep neural networks over-fit to the identities of the subjects present in the dataset. Due to this over-fitting phenomenon, large portion of the representation used for facial expression recognition contains identity-related information as opposed to capturing expression features, as reported in \cite{r1}, \cite{r2}, \cite{r8}. As a result, the performance of facial expression recognition degrades on unseen subjects during testing. 

Apart from the problem of inter-subject variation, another major problem that arises in real-time automatic facial expression recognition is that most datasets are compiled in a lab controlled environment where six posed facial expressions exhibited by the subjects are captured. These six basic expressions are anger, disgust, sadness, happiness, fear and surprise. It is therefore assumed that the captured posed expression images correspond to pure forms of these six basic expressions. However, it has been reported in the literature that humans are able to display a wide range of facial expressions \cite{r37}. For example in \cite{r38} it has been studied that up to 7000 combinations of Action Units (AUs) are exhibited by humans in everyday life. Therefore, many of those images in FER datasets do not contain pure facial expressions which makes it difficult even for humans to classify them correctly.

\begin{figure*}[ht!]
\centering
%[scale=1, width=.01\textwidth]
\includegraphics[width=16cm,, height=7.5cm]{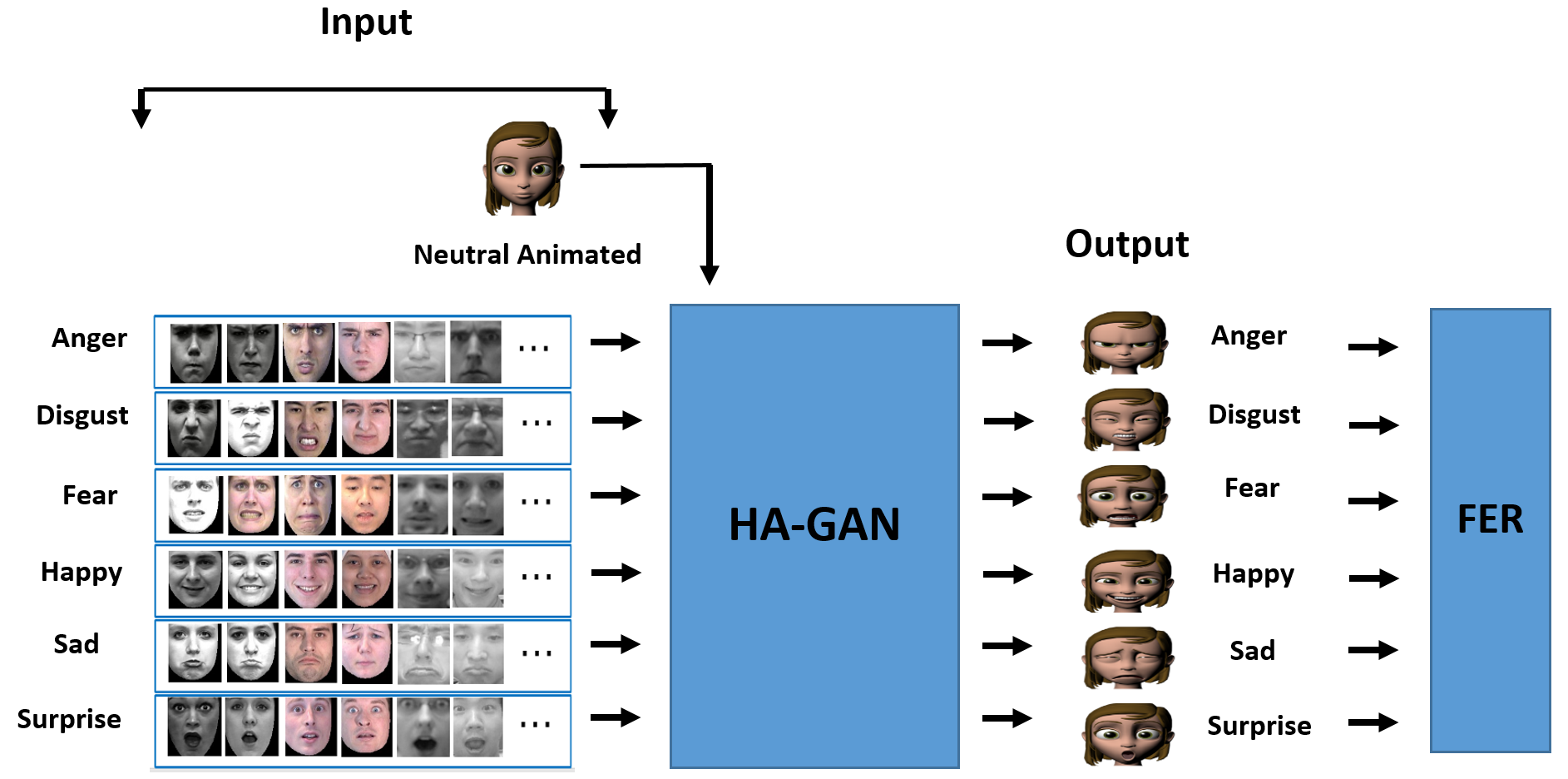}
\caption{Overall framework of our HA-GAN}
\label{fig:1}
\end{figure*}

In order to overcome the above mentioned problems, we propose a novel Human-to-Animated conditional Generative Adversarial Network (HA-GAN) to reduce inter-subject variations by transferring facial expression from input human image to an animated image having a pure facial expression. The overall framework of our proposed HA-GAN is shown in Figure \ref{fig:1}. Specifically, we train our HA-GAN to generate animated expression images using the Facial Expression Research Group-Database (FERGDB) in which the expression images are diligently and meticulously designed by expert animators and artists to have pure facial expressions. Our HA-GAN learns the human-expression-to-animated-expression mapping function and, after training, generates realistically looking animated expression images. We argue that the facial expressions in these generated animated images are pure and uniform. Inspired by the experimental results reported in \cite{r39}, where it has been observed that the accuracy of FER is about $4\%$ higher in case of animated images than with human images, we perform FER on the generated animated expression images having pure facial expressions with a fixed identity.

There are many exciting applications of our proposed HA-GAN, such as it can be used to predict the 3D rig parameters of an animated character by first mapping 2D human expression images to 2D character expression images. Then, since each 2D character image is rendered from its 3D facial rig, we can learn a mapping to predict the 3D rig parameters without leveraging any dataset having the mapping from 2D human expression iamges to 3D character rig parameters. In \cite{r44}, a joint embedding is learned using a multi-training technique in which human and character expressions are mapped based on the distance between human and animated expression features, and thus the corresponding animated character is retrieved. The 3D rig parameters of the retrieved animated character is then calculated. While our proposed technique can be used to retrieve the corresponding character by generating the its image employing our trained generator. 

The major contributions of this paper are summarized as follows:

\begin{itemize}
\item We present a novel HA-GAN which learns a many-to-one expression mapping function to transfer expressions from human images to a fixed animated identity in order to remove inter-subject variations. 
\item To the best of our knowledge, this is the first work which is aimed to address the problem of impure and non-uniform expressions exhibited by human subjects in FER datasets. In order to overcome this problem we train our end-to-end HA-GAN framework to generate realistic looking animated expression images with pure facial expressions, which are then used for FER.
\item There are many exciting applications of our proposed HA-GAN, such as it can be used to predict the 3D rig parameters of animated characters from 2D human expression images.
\item Initial experimental results show that the proposed technique produces comparable results to the state-of-the-art methods.
\end{itemize}

\section{Related Work}

Facial expression recognition is one of the most widely studied topics in the computer vision research community over the past decades \cite{r6} \cite{r7} \cite{r8}. The main objective of FER is to learn features that contain mostly expression-related information that is discriminative and invariant to confounds such as pose, illumination, and identity-related information (age, race, gender etc). 

Many feature extraction techniques have been proposed by researchers in the past, and these techniques can be divided into two main categories: human-crafted features and automatically-learned features. Most human-crafted feature extraction techniques were developed before the deep learning era, such as Histograms of Oriented Gradients (HOG) \cite{r9}, \cite{r27}, Scale Invariant Feature Transform (SIFT) features \cite{r10}, \cite{r11}, histograms of Local Binary Patterns (LBP) \cite{r26}, \cite{r12}, \cite{r13}, and histograms of Local Phase Quantization (LPQ) \cite{r14} \cite{r31}. These techniques can be further categorized into static and dynamic techniques, where in static methods FER representations are learned from images, and in dynamic methods FER features are extracted from sequence of images.

Both hand-crafted feature extraction techniques and automatic FER representation learning methods have been applied to overcome one of the major problems in FER: to reduce inter-subject variations. Although human-crafted features produce acceptable results in lab controlled environments where the expression images are captured in a constant illumination and stable head pose. However, these features fail on spontaneous and real time data with varying illumination and head position. With the recent success of deep CNN, automatic feature learning for FER has been extensively studied by computer vision researchers and various techniques such as \cite{r15}, \cite{r16}, \cite{r17} \cite{r28} \cite{r30} \cite{r32} \cite{r33} have been developed to increase the robustness of FER. However, the performance of these automatically learned deep representations are affected by large variations in identity related facial characteristics such as age, ethnicity, gender, etc of subjects involved in the dataset. As a result, the generalization capability of the trained model is negatively affected by this over-fitting phenomena and thus the FER accuracy is degraded on unseen subjects. Therefore, despite all the progress in improving the generalizability of FER, the main problem of eliminating the negative effect of inter-subject variations on FER is still an open challenge for the research community. 

To improve the discriminative property of extracted features various techniques \cite{r18}\cite{r19} have been developed in the past that are aimed to reduce intra-class variations and increase inter-class differences. Identity-Aware CNN (IACNN) \cite{r4} is a recently proposed technique to reduce the effect of identity related features by applying an expression-sensitive contrastive loss and an identity-sensitive contrastive loss. However, the performance of FER is negatively influenced due to large data expansion caused as a result of compilation of training data in the form of image pairs \cite{r2}. Similarly in \cite{r5}, person-independent expression features are extracted by employing a De-expression Residue Learning (DeRL) method using conditional GANs. In DeRL, the cGAN is used to synthesize a neutral image from an expression image by employing an encoder-decoder based generator. The learned subject-independent expression feature is then extracted from intermediate layers of both the encoder and decoder parts of the generator for FER. However, the cGAN based training followed by the joint training process of extracted expression features from intermediate layers makes the proposed DeRL method computationally very costly. Likewise, in \cite{r43}, a disentangled expression representation method is proposed in which an encoder-decoder based DE-GAN is used to disentangle expression features from identity information. In another attempt to reduce the effect of identity-related features, in \cite{r2}, an Identity-free conditional Generative Adversarial Network (IF-GAN) method is proposed to synthesize a common synthetic image by transferring the facial expression information from the input image to the synthesized image. FER is then carried out using the generated synthetic expression images in order to mitigate the effect of inter-subject variations in data. Although, the effect of individual variations of subjects are reduced in the IF-GAN method, the accuracy of FER, however, depends on the quality of transferred expressions, and since the expressions are transferred on a human image, the problem of impurity of human expressions still persists. Therefore, to the best of our knowledge, our proposed HA-GAN based method is the first work to address both the inter-subject variation problem and the issue of impure and non-uniform human expressions in FER.

\section{Proposed Method}
GANs are deep generative networks that learn a mapping function to generate realistic images $y$ from random noise vector $z$, $G: z\,\to\,y$ \cite{r40}. In order to achieve this task two neural networks: a generator, $G$ and a discriminator $D$ are trained in an adversarial manner. The main objective of the generator is to model the probability distribution of the data, while the job of the discriminator is to find the probability of a sample being taken from the real data distribution or from the fake data distribution of samples generated by the generator. Thus, a generator $G$ is trained to generate realistic samples from noise vector $z$ to fool discriminator $D$ to classify it as a real sample.
The objective function of GANs are formulated as follows:
\begin{align}
\underset{G}{\mathrm{min}}\underset{D}{\mathrm{max}}{L}(D,G) =E[\log(D(y))+
&E[\log(1-D(G(z)))]\notag\\
\end{align}
where $y$ corresponds to a sample drawn from the  real data distribution and $z$ denotes a random noise vector taken from noise distribution $p_z$.

In contrast, Conditional GANs (cGANs) learn a mapping from random noise vector $z$ to output $y$, while being conditioned on sample $x$, $G: \{x,z\}\,\to\,y$ \cite{r41}. The objective function of a cGAN can be formulated as follows:
\begin{align}
\underset{G}{\mathrm{min}}\underset{D}{\mathrm{max}}{L}(D,G) =E[\log(D(x,y))+\notag\\
E[\log(1-D(x,G(x,z)))]\notag\\
\end{align}
Here, the main goal of $G$ is to minimize this objective function, while discriminator $D$ tries to maximize it in an adversarial manner. 

\subsection{The Proposed HA-GAN}

The main goal of our novel HA-GAN is to encode the expression information from an input image and learn to transfer that encoded expression information to an animated generated image. Facial expression recognition is then performed on the generated animated expression images.

\begin{figure*}
\begin{center}
%[scale=1, width=.01\textwidth]
\includegraphics[width=16.5cm,, height=6.5cm]{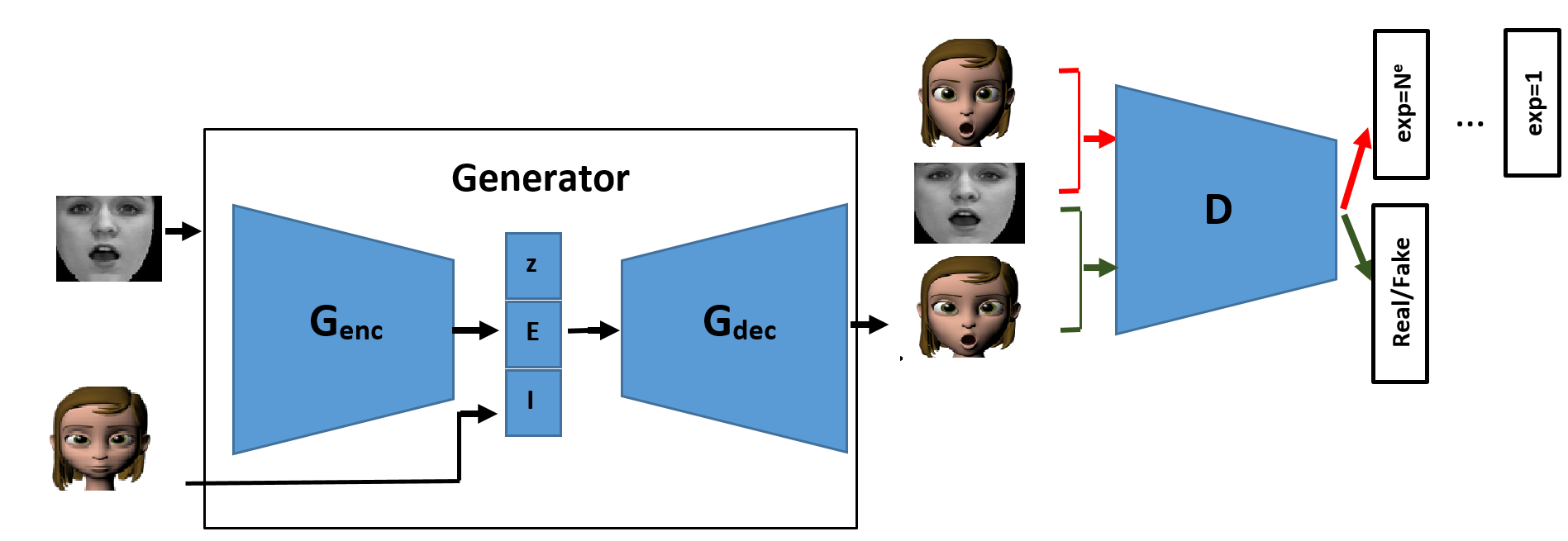}
\end{center}
   \caption{The complete architecture of HA-GAN: an encoder-decoder based generator $G$, a discriminator $D$, with two parts, one for real and fake classification, while the second part of $D$ is for classification of expressions.}
\label{fig:2}
\end{figure*}

% ***********************************

\subsubsection{Generator: }

The input to the generator of our HA-GAN is a human expression image  $I_{HE}$ with an expression $E$, a conditional animated character information in the form of one-hot vector and a random noise vector $z$. The conditional animated character information corresponds to the identity of the character having a neutral expression $N$ and it is denoted by $Id_{AN}$. Random noise $z$ is used to model other variations such as head-pose, etc. Given these inputs, the main goal of our generator $G$ is to encode the expression information $E$ from the input human image $I_{HE}$, and generate an animated character image $\bar I_{AE}$,  exhibiting the encoded expression $E$ present in the input human expression image. In order to achieve this task, we have used an encoder $G_E$ and a decoder $G_D$ based generator. The main objective of our encoder $G_E$ is to encode the expression information $E$ from the input image $I_{HE}$. This encoded expression information $E$ is then combined with the identity information $Id_{AN}$ and noise vector  $z$, and fed to the decoder $G_D$ part of our generator to generate an animated character image $\bar I_{AE}$ with the same expression as the input image. The main objective of $G$ is to generate realistic looking character expression image:  $G(I_{HE},Id_{AN},z) = \bar I_{AE}$  to fool $D$ to classify the expression of $\bar I_{AE}$ to the expression in the input image $I_{HE}$ using the following objective function: 
\begin{align}
\underset{G}{\mathrm{max}}\mathcal{L}(D,G) ={}&\underset{z\sim p_z(z),{Id}\sim p_{Id}({Id})}{\mathrm{E_{I_{HE},E\sim p_d(I_{HE},E)}}}[\log({D_E^e}{(G(I_{HE},Id_{AN},z))}\notag\\
&+\log({D^{R}}{(I_{HE},G(I_{HE},Id_{AN},z))}]\notag\\
\end{align} 

Inspired by the previous GAN optimization techniques such as in \cite{r42}, we add L1 distance metrics in the above mentioned GAN loss. But in our case instead of calculating the L1 distance between the generated image and input image as in \cite{r42}, we use L1 distance metric between the generated character image $\bar I_{AE}$ and the neutral character image $I_{AN}$ of the identity specified by $Id_{AN}$. We use L1 distance metric between $\bar I_{AE}$ and $I_{AN}$ instead due to the fact that our input image comes from a different domain (human face images) and our generated output image belongs to animated character images. Therefore, we have experimentally observed that the quality of the generated character images are improved significantly by training the generator to generate fake images close to the identity of ground truth character images by minimizing the L1 distance between the two images: 
\begin{align}
\mathcal{L}_{L_1} (G)=E[\Vert I_{AN} - G(I_{AN},I_{HE})\Vert_1]\notag\\
\end{align}
We also experimented using L2 distance, but we observed that the quality of images generated using L1 distance is less blurred as compared to using L2 distance.
\subsubsection{Discriminator: } In contrast to a traditional GAN, where the task of the discriminator is to identify real and fake images, the discriminator $D$ in HA-GAN is a multi-task convolutional neural network having two objectives: 1. to classify between real and fake images, and 2. to recognize facial expressions. Thus our discriminator $D$ has two parts, i.e $D = [D^e, D^R]$, where $D^e \in R^{N^e}$ is for the classification of expressions i.e $N^e$ corresponds to six basic expressions and $D^R$ is for the classification of real and fake images. The objective function of our discriminator $D$ is given by:
\begin{align}
\underset{G}{\mathrm{max}}\mathcal{L}(D,G) ={}&{\mathrm{E_{I_{HE},E\sim p_d(I_{HE},E)}}}[\log({D_E^e}{(I_{HE}, I_{AE})]}\notag\\
&+\underset{z\sim p_z(z),{Id}\sim p_{Id}({Id})}{\mathrm{E_{I_{HE},E\sim p_d(I_{HE},E)}}}[\log({D^{R}}{(I_{HE},\bar I_{AE})}]\notag\\
\end{align}

The first part of the above equation corresponds to the objective of the discriminator $D^e$ to maximize the probability of classifying the expressions of the paired images: $(I_{HE}, I_{AE})$ to its ground truth expression label, $E$. While the second part of the equation is for $D^R$ to maximize the classification probability of $\bar I_{AE}$ according to  its true label, i.e fake. 

In order to reduce computation cost, the discriminator $D$ is designed in such a way that the two parts share initial down sampling convolutional layers, i.e 4 CNN blocks with 16, 32, 64, 128 channels and a FC layer to generate a 1024-dimensional vector. Then it is divided into two branches, one for $D^R$ and the other for $D^e$. The $D^R$ branch has another FC layer which is then connected to the output neuron, while the $D^e$ leg has two FC layers and then an output layer for expression classification.

\section{Facial Expression Recognition}
After the end-to-end training of our HA-GAN, during testing, the $D^R$ leg of discriminator $D$ is discarded, and we employ only the trained $D^e$ branch of $D$ for facial expression recognition based on the realistic looking generated animated character images having pure transferred facial expression, as shown in Figure \ref{fig:2}.

\section{Experiments}
To validate our hypothesis and to illustrate the effectiveness of our novel HA-GAN based FER technique, we performed experiments on benchmark datasets: CK+ \cite{r20}, Oulu-CASIA \cite{r21} and MMI \cite{r22} while training our generator using Facial Expression Research Group-Database (FERGDB) \cite{r39} to generate synthetic character expression images.

\subsection{Implementation Details}
Face detection and face alignment is performed based on the facial landmarks obtained by employing Convolutional Experts Constrained Local Model (CE-CLM) \cite{r23}. Afterwards, $75 \times 75$ patches are randomly cropped from the aligned face images. Due to small dataset size, data augmentation is applied to increase the size of datasets in order to avoid the over-fitting issue. Five samples of size $75 \times 75$ are cropped-out from five locations: four corners and the center position of each image, and each of those $75 \times 75$ cropped samples are then rotated at ten angles i.e $-150^\circ$, $-120^\circ$, $-90^\circ$, $-60^\circ$, $-30^\circ$, $30^\circ$, $60^\circ$, $90^\circ$, $120^\circ$, $150^\circ$. The rotated images are then horizontally flipped to further increase the number of training data. As a result of this data augmentation process, the size of the training data is increased by 110 times. The data augmentation process is only applied to human expression images. 

We followed the optimization strategies applied in \cite{r3} to optimize our networks. The network parameters of both $G$ and $D$ are updated during back propagation one by one. Following \cite{r3}, $G$ is trained to maximize $\log{D_{E}^e}(G(I_{AN},I_{HE})$ rather than training $G$ to minimize $\log(1-{D_{E}^e}(G(I_{AN},I_{HE}))$. To reduce the learning rate of $D$ relative to $G$, the objective function is divided by two during the optimization of $D$. Adam optimizer \cite{r25} produced better results with a batch size of 130, learning rate of 0.0002 and momentum of 0.5. We trained our HA-GAN framework for 200 epochs using NVIDIA TESLA V100 GPU. 

\subsection{Experimental Results}
The \textbf{Facial Expression Research Group-Database (FERGDB)} \cite{r39} is used in our project to learn the mapping of human expressions from human images to animated character images. The FERG database consists of six stylized characters (identities), each of which is designed and created by animators to have images of six basic expression along with neutral images. One of these six animated characters is used in our proposed method to transfer an expression from an input human image to the animated character image. 

The \textbf{Extended Cohn-Kanade database CK+} \cite{r20} is a popular facial expression recognition benchmark that contains 327 videos sequences from 118 subjects. Each of these video sequences are labelled as one of the seven expressions, i.e. anger, contempt, disgust, fear, happiness, sadness, and surprise. The video sequences start with a neutral expression image and end at a peak expression. To create the training and testing data, the last three frames of each sequence are selected as an expression image. For validation purposes, the dataset is divided into training and testing subsets in an identity-independent manner. 

Our initial result, which is the average accuracy of two runs of our 10 fold validation process is shown in Table \ref{table:1}. As it can be seen in Table \ref{table:1} that our accuracy is higher than most of state-of-the-art techniques. It is also worth noting that our technique is based on images and we are not using the entire video sequences unlike many of the techniques in Table \ref{table:1}. We are also not using any transfer learning technique unlike \cite{r5} due to limited time and computational resources, which, we believe, if incorporated in our technique will significantly boost our expression transfer and recognition accuracy.

\begin{table}
\begin{center}
\begin{tabular}{|l| c| c|} 
 %\hline
 %\multicolumn{4}{|c|}{Country List} \\
 \hline
 Method&Setting/Classes&Acc\\
 \hline
 LBP-TOP\cite{r26}   & Dynamic/7    &88.99\\
 HOG 3D\cite{r27}&   Dynamic/7  & 91.44\\
 3DCNN \cite{r28} &Dynamic/7 & 85.9\\
 STM-Explet\cite{r29}    &Dynamic/7 & 94.19\\
 IACNN\cite{r4}&   Static/7  & 95.37\\
 DTAGN\cite{r30}&   Static/7  & 97.25\\
 DeRL\cite{r5}&   Static/7  & 97.30\\
 \hline
 CNN(baseline)& Static/6  & 90.34\\
 \textbf{HA-GAN(Ours)}& Static/6  & \textbf{96.14}\\
 \hline
\end{tabular}
\end{center}
\caption{CK+: Accuracy for six expressions classification.}
\label{table:1}
\end{table}

The \textbf{MMI dataset} \cite{r22} is one of the most challenging facial expression database due to two major reasons: 1. it is a small dataset containing only 236 video image sequences corresponding to six facial expressions of 31 subjects. 2. the inter-personal variations in this dataset is large because the same facial expression is performed differently by different identities. Similarly many of the subjects involved in this dataset wear accessories (e.g., glasses, cap, scarf, mustache). The training and testing dataset is compiled using only the frontal view images from 208 sequences from 31 subjects. The three middle frames in each sequence corresponds to the peak expression and thus they are selected in our training and testing data. An identity-independent partitioning of the data is made to carry out cross validation. 

The average accuracy of two run of our 10-fold validation process is shown in Table \ref{table:2}. As it can be seen that the accuracy of our technique is much higher than the accuracy of our CNN baseline network. Comparing our results with static techniques, as it can be seen that we are quite close to DeRL\cite{r5} methods accuracy. Although STM-Explet\cite{r29} shows the highest recognition accuracy their technique is a dynamic technique where the information is extracted from video sequences, which is not always available in realtime applications such as image based FER.     

 The \textbf{Oulu-CASIA dataset} \cite{r21} is divided into three parts based on the images obtained in three different lighting environments with two different cameras. During this project, the images captured using the VIS camera under strong illumination conditions are used for training and testing. The Oulu-CASIA VIS data contains 480 sequences from 80 subjects, where each sequence is labeled as one of the six basic expressions. Each video sequence starts with a neutral expression image and ends at a peak expression. The last three images in each sequence are taken to compile our training and testing datasets. This dataset is then divided into disjoint subsets based on identity to perform cross-validation.
 
 The average accuracy of two run of our 10-fold validation process is shown in Table \ref{table:3}. Note that LBP-TOP\cite{r26}, HOG 3D\cite{r27}, STM-Explet\cite{r29} and Atlases\cite{r31} use the temporal information of video sequences, and our HA-GAN result is much higher than their results without exploiting any temporal information. 

\begin{table}
\begin{center}
\begin{tabular}{|l| c| c|} 
 %\hline
 %\multicolumn{4}{|c|}{Country List} \\
 \hline
 Method&Setting&Accuracy\\
 \hline
 LBP-TOP\cite{r26}   & Dynamic    &59.51\\
 HOG 3D\cite{r27}&   Dynamic  & 60.89\\
 STM-Explet\cite{r29}    &Dynamic & 75.12\\
 IACNN\cite{r4}&   Static  & 71.55\\
 DTAGN\cite{r30}&   Static  & 70.24\\
 DeRL\cite{r5}&   Static  & 73.23\\
 \hline
 CNN(baseline)& Static  & 58.46\\
 \textbf{HA-GAN(Ours)}& Static  & \textbf{71.87}\\
 \hline
\end{tabular}
\end{center}
\caption{MMI: Accuracy for six expressions classification.}
\label{table:2}
\end{table}

\begin{table}
\begin{center}
\begin{tabular}{|l| c| c|} 
 %\hline
 %\multicolumn{4}{|c|}{Country List} \\
 \hline
 Method&Setting&Accuracy\\
 \hline
 LBP-TOP\cite{r26}   & Dynamic    &68.13\\
 HOG 3D\cite{r27}&   Dynamic  & 70.63\\
 STM-Explet\cite{r29}    &Dynamic & 74.59\\
 Atlases\cite{r31}&   Dynamic  & 75.52\\
 FN2EN\cite{r32}&   Static  & 87.71\\
 PPDN\cite{r33}&   Static  & 84.59\\
 DTAGN\cite{r30}&   Static  & 81.46\\
 DeRL\cite{r5}&   Static  & 88.0\\
 \hline
 CNN(baseline)& Static  & 73.14\\
 \textbf{HA-GAN(Ours)}& Static  & \textbf{88.26}\\
 \hline
\end{tabular}
\end{center}
\caption{Oulu-CASIA: Accuracy for six expressions classification.}
\label{table:3}
\end{table}

\section{Conclusions}
In this paper we have presented a novel HA-GAN framework that is trained in an end-to-end manner to overcome two major problems in automatic facial expression recognition. The first problem that we address in this work is to eliminate the inter-subject variations present in facial expression datasets. Using our HA-GAN architecture we transfer the expression information from an input human image to an animated character image having a fixed identity. Our initial experimental results illustrate that learning this many-to-one mapping can help to overcome the problem of inter-subject variations. The second challenging issue in FER that we tackle is the impurity of posed human expressions. Due to impure and non-uniform exhibition of human facial expressions, the performance of facial expression algorithms are largely degraded, especially in real world settings. In this project we train our HA-GAN model to transfer human expressions to an animated character having pure expressions that were carefully designed by expert animators. Inspired by the previous research findings in which it is found that the accuracy of FER is higher in case of animated images as compared to FER accuracy on human images, we then use the generated animated character expression images for the purpose of facial expression recognition. Our experimental results show that the proposed method produces comparable or even better results than state-of-the-art techniques.

%-------------------------------------------------------------------------

{\small
\bibliographystyle{ieee}
\bibliography{egbib}
}

\end{document}